# Model Adaptation via Model Interpolation and Boosting for Web Search Ranking

**Jianfeng Gao[*], Qiang Wu[*], Chris Burges[*], Krysta Svore[*],**
**Yi Su[#], Nazan Khan[$], Shalin Shah[$], Hongyan Zhou[$]**

[*]Microsoft Research, Redmond, USA
{jfgao; qiangwu; cburges; ksvore}@microsoft.com
[#]Johns Hopkins University, USA
suy@jhu.edu
[$]Microsoft Bing Search, Redmond, USA
{nazanka; a-shas; honzhou}@microsoft.com

## Abstract

This paper explores two classes of model adaptation methods for Web search ranking: Model Interpolation and error-driven learning approaches based on a boosting algorithm. The results show that model interpolation, though simple, achieves the best results on all the open test sets where the test data is very different from the training data. The tree-based boosting algorithm achieves the best performance on most of the closed test sets where the test data and the training data are similar, but its performance drops significantly on the open test sets due to the instability of trees. Several methods are explored to improve the robustness of the algorithm, with limited success.

## 1 Introduction

We consider the task of ranking Web search results, i.e., a set of retrieved Web documents (URLs) are ordered by relevance to a query issued by a user. In this paper we assume that the task is performed using a *ranking model* (also called *ranker* for short) that is learned on labeled training data (e.g., human-judged query-document pairs). The ranking model acts as a function that maps the feature vector of a query-document pair to a real-valued score of relevance.

Recent research shows that such a learned ranker is superior to classical retrieval models in two aspects (Burges et al., 2005; 2006; Gao et al., 2005). First, the ranking model can use arbitrary features. Both traditional criteria such as TF-IDF and BM25, and non-traditional features such as hyperlinks can be incorporated as features in the ranker. Second, if large amounts of high-quality human-judged query-document pairs were available for model training, the ranker could achieve significantly better retrieval results than the traditional retrieval models that cannot benefit from training data effectively. However, such training data is not always available for many search domains, such as non-English search markets or person name search.

One of the most widely used strategies to remedy this problem is model adaptation, which attempts to adjust the parameters and/or structure of a model trained on one domain (called the *background* domain), for which large amounts of training data are available, to a different domain (the *adaptation* domain), for which only small amounts of training data are available. In Web search applications, domains can be defined by query types (e.g., person name queries), or languages, etc.

In this paper we investigate two classes of model adaptation methods for Web search ranking: Model Interpolation approaches and error-driven learning approaches. In model interpolation approaches, the adaptation data is used to derive a domain-specific model (also called in-domain model), which is then combined with the background model trained on the background data. This appealingly simple concept provides fertile ground for experimentation, depending on the level at which the combination is implemented (Bellegarda, 2004). In error-driven learning approaches, the background model is adjusted so as to minimize the ranking errors the model makes on the adaptation data (Bacchiani et al., 2004; Gao et al. 2006). This is arguably more powerful than model interpolation for two reasons. First, by defining a proper error function, the method can optimize more directly the measure used to assess the final quality of the Web search system, e.g., *Normalized Discounted Cumulative Gain* (Javelin & Kekalainen, 2000) in this study. Second, in this framework, the model can be adjusted to be as fine-grained as necessary. In this study we developed a set of error-driven learning methods based on a boosting algorithm where, in an incremental manner, not only each feature weight could be changed separately, but new features could be constructed.

We focus our experiments on the robustness of the adaptation methods. A model is robust if it performs reasonably well on unseen test data that could be significantly different from training data. Robustness is important in Web search applications. Labeling training data takes time. As a result of the dynamic nature of Web, by the time the ranker is trained and deployed, the training data may be more or less out of date. Our results show that the model interpolation is much more robust than the boosting-based methods. We then explore several methods to improve the robustness of the methods, including regularization, randomization, and using shallow trees, with limited success.

## 2 Ranking Model and Quality Measure in Web Search

This section reviews briefly a particular example of rankers, called LambdaRank (Burges et al., 2006), which serves as the baseline ranker in our study.

Assume that training data is a set of input/output pairs (**x**, $y$). **x** is a feature vector extracted from a query-document pair. We use approximately 400 features, including dynamic ranking features such as term frequency and BM25, and statistic ranking features such as PageRank. $y$ is a human-judged relevance score, 0 to 4, with 4 as the most relevant.

LambdaRank is a neural net ranker that maps a feature vector **x** to a real value $y$ that indicates the relevance of the document given the query (relevance score). For example, a linear LambdaRank simply maps **x** to $y$ with a learned weight vector **w** such that $y = \mathbf{w} \cdot \mathbf{x}$. (We used nonlinear LambdaRank in our experiments). LambdaRank is particularly interesting to us due to the way **w** is learned. Typically, **w** is optimized w.r.t. a cost function using numerical methods if the cost function is smooth and its gradient w.r.t. **w** can be computed easily. In order for the ranker to achieve the best performance in document retrieval, the cost function used in training should be the same as, or as close as possible to, the measure used to assess the quality of the system. In Web search, *Normalized Discounted Cumulative Gain* (NDCG) (Jarvelin and Kekalainen, 2000) is widely used as quality measure. For a query, NDCG is computed as

$$\mathcal{N}_i = N_i \sum_{j=1}^{L} \frac{2^{r(j)} - 1}{\log(1+j)}, \quad (1)$$

where $r(j)$ is the relevance level of the $j$-th document, and the normalization constant $N_i$ is chosen so that a perfect ordering would result in $\mathcal{N}_i = 1$. Here $L$ is the ranking truncation level at which NDCG is computed. The $\mathcal{N}_i$ are then averaged over a query set. However, NDCG, if it were to be used as a cost function, is either flat or discontinuous everywhere, and thus presents challenges to most optimization approaches that require the computation of the gradient of the cost function.

LambdaRank solves the problem by using an implicit cost function whose gradients are specified by rules. These rules are called *λ-functions*. Burges et al. (2006) studied several λ-functions that were designed with the NDCG cost function in mind. They showed that LambdaRank with the best λ-function outperforms significantly a similar neural net ranker, RankNet (Burges et al., 2005), whose parameters are optimized using the cost function based on cross-entropy.

The superiority of LambdaRank illustrates the key idea based on which we develop the model adaptation methods. We should always adapt the ranking models in such a way that the NDCG can be optimized as directly as possible.

## 3 Model Interpolation

One of the simplest model interpolation methods is to combine an in-domain model with a background model at the model level via linear interpolation. In practice we could combine more than two in-domain/background models. Letting *Score*($q$, $d$) be a ranking model that maps a query-document pair to a relevance score, the general form of the interpolation model is

$$Score(q,d) = \sum_{i=1}^{N} \alpha_i Score_i(q,d), \quad (2)$$

where the $\alpha's$ are interpolation weights, optimized on validation data with respect to a predefined objective, which is NDCG in our case. As mentioned in Section 2, NDCG is not easy to optimize, for which we resort to two solutions, both of which achieve similar results in our experiments.

The first solution is to view the interpolation model of Equation (2) as a linear neural net ranker where each component model *Score$_i$*(.) is defined as a feature function. Then, we can use the Lambda-Rank algorithm described in Section 2 to find the optimal weights.

An alternative solution is to view interpolation weight estimation as a multi-dimensional optimization problem, with each model as a dimension. Since NCDG is not differentiable, we tried in our experiments the numerical algorithms that do not require the computation of gradient. Among the best performers is the Powell Search algorithm (Press et al., 1992). It first constructs a set of $N$ virtual directions that are conjugate (i.e., independent with each other), then it uses *line search N* times, each on one virtual direction, to find the optimum. Line search is a one-dimensional optimization algorithm. Our implementation follows the one described in Gao et al. (2005), which is used to optimize the averaged precision.

The performance of model interpolation depends to a large degree upon the quality and the size of adaptation data. First of all, the adaptation data has to be "rich" enough to suitably characterize the new domain. This can only be achieved by collecting more in-domain data. Second, once the domain has been characterized, the adaptation data has to be "large" enough to have a model reliably trained. For this, we developed a method, which attempts to augment adaptation data by gathering similar data from background data sets.

The method is based on the *k-nearest-neighbor* (kNN) algorithm, and is inspired by Bishop (1995). We use the small in-domain data set *D1* as a seed, and expand it using the large background data set *D2*. When the relevance labels are assigned by humans, it is reasonable to assume that queries with the lowest information entropy of labels are the least noisy. That is, for such a query most of the URLs are labeled as highly relevant/not relevant documents rather than as moderately relevance/not relevant documents.

Due to computational limitations of kNN-based algorithms, a small subset of queries from *D1* which are least noisy are selected. This data set is called *S1*. For each sample in *D2*, its 3-nearest neighbors in *S1* are found using a cosine-similarity metric. If the three neighbors are within a very small distance from the sample in *D2*, and one of the labels of the nearest neighbors matches exactly, the training sample is selected and is added to the expanded set *E2*, in its own query. This way, *S1* is used to choose training data from *D2*, which are found to be close in some space.

---

1    Set $F_0(\mathbf{x})$ be the background ranking model
2    **for** $m = 1$ **to** $M$ **do**
3      $y_i' = -\left[\frac{\partial L(y_i, F(\mathbf{x}_i))}{\partial F(\mathbf{x}_i)}\right]'_{F(\mathbf{x})=F_{m-1}(\mathbf{x})}$, for $i = 1... N$
4      $(h_m, \beta_m) = \underset{h,\beta}{\operatorname{argmin}} \sum_{i=1}^{N} [y_i' - \beta h(\mathbf{x}_i)]^2$
5      $F_m(\mathbf{x}) = F_{m-1}(\mathbf{x}) + \beta_m h(\mathbf{x})$

**Figure 1.** The generic boosting algorithm for model adaptation.

This process effectively creates several data points in close neighborhood of the points in the original small data set *D1*, thus expanding the set, by jittering each training sample a little. This is equivalent to training with noise (Bishop, 1995), except that the training samples used are actual queries judged by a human. This is found to increase the NDCG in our experiments.

## 4    Error-Driven Learning

Our error-drive learning approaches to ranking modeling adaptation are based on the Stochastic Gradient Boosting algorithm (or the boosting algorithm for short) described in Friedman (1999). Below, we follow the notations in Friedman (2001).

Let adaptation data (also called *training data* in this section) be a set of input/output pairs {$\mathbf{x}_i, y_i$}, $i = 1…N$. In error-driven learning approaches, model adaptation is performed by adjusting the background model into a new in-domain model $F: x \rightarrow y$ that minimizes a loss function $L(y, F(\mathbf{x}))$ over all samples in training data

$$F^* = \underset{F}{\operatorname{argmin}} \sum_{i=1}^{N} L(y_i, F(\mathbf{x}_i)). \quad (3)$$

We further assume that $F(\mathbf{x})$ takes the form of additive expansion as

$$F(\mathbf{x}) = \sum_{m=0}^{M} \beta_m h(\mathbf{x}; \mathbf{a}_m), \quad (4)$$

where $h(\mathbf{x}; \mathbf{a})$ is called *basis function*, and is usually a simple parameterized function of the input $\mathbf{x}$, characterized by parameters $\mathbf{a}$. In what follows, we drop $\mathbf{a}$, and use $h(\mathbf{x})$ for short. In practice, the form of $h$ has to be restricted to a specific function family to allow for a practically efficient procedure of model adaptation. $\beta$ is a real-valued coefficient.

Figure 1 is the generic algorithm. It starts with a base model $F_0$, which is a background model. Then for $m = 1, 2, …, M$, the algorithm takes three steps to adapt the base model so as to best fit the adaptation data: (1) compute the residual of the current base model (line 3), (2) select the optimal basis function (line 4) that best fits the residual, and (3) update the base model by adding the optimal basis function (line 5). The two model adaptation algorithms that will be described below follow the same 3-step adaptation procedure. They only differ in the choice of $h$. In the LambdaBoost algorithm (Section 4.1) $h$ is defined as a single feature, and in LambdaSMART (Section 4.2), $h$ is a regression tree.

| | |
|---|---|
| 1 | Set $F_0(\mathbf{x})$ to be the background ranking model |
| 2 | **for** $m$ = 1 **to** $M$ **do** |
| 3 | compute residuals according to Equation (5) |
| 4 | select best $h_m$ (with its best $\beta_m$), according to *LS*, computed by Equations (8) and (9) |
| 5 | $F_m(\mathbf{x}) = F_{m-1}(\mathbf{x}) + \upsilon \beta_m h(\mathbf{x})$ |

**Figure 2.** The LambdaBoost algorithm for model adaptation.

Now, we describe the way residual is computed, the step that is identical in both algorithms. Intuitively, the residual, denoted by $y'$ (line 3 in Figure 1), measures the amount of errors (or loss) the base model makes on the training samples. If the loss function in Equation (3) is differentiable, the residual can be computed easily as the negative gradient of the loss function. As discussed in Section 2, we want to directly optimize the NDCD, whose gradient is approximated via the $\lambda$-function. Following Burges et al. (2006), the gradient of a training sample $(\mathbf{x}_i, y_i)$, where $\mathbf{x}_i$ is a feature vector representing the query-document pair $(q_i, d_i)$, w.r.t. the current base model is computed by marginalizing the $\lambda$-functions of all document pairs, $(d_i, d_j)$, of the query, $q_i$, as

$$y_i' = \sum_{j \neq i} \Delta\text{NDCG} \cdot \frac{\partial C_{ij}}{\partial o_{ij}}, \quad (5)$$

where $\Delta$NDCG is the NDCG gained by swapping those two documents (after sorting all documents by their current scores); $o_{ij} \equiv s_i - s_j$ is the difference in ranking scores of $d_i$ and $d_j$ given $q_i$; and $C_{ij}$ is the cross entropy cost defined as

$$C_{ij} \equiv C(o_{ij}) = s_j - s_i + \log(1 + \exp(s_i - s_j)). \quad (6)$$

Thus, we have

$$\frac{\partial C_{ij}}{\partial o_{ij}} = \frac{-1}{1 + \exp(o_{ij})}. \quad (7)$$

This $\lambda$-function essentially uses the cross entropy cost to smooth the change in NDCG obtained by swapping the two documents. A key intuition behind the $\lambda$-function is the observation that NDCG does not treat all pairs equally; for example, it costs more to incorrectly order a pair, where the irrelevant document is ranked higher than a highly relevant document, than it does to swap a moderately relevant/not relevant pair.

### 4.1 The LambdaBoost Algorithm

In LambdaBoost, the basis function $h$ is defined as a single feature (i.e., an element feature in the feature vector $\mathbf{x}$). The algorithm is summarized in Figure 2. It iteratively adapts a background model to training data using the 3-step procedure, as in Figure 1. Step 1 (line 3 in Figure 2) has been described.

Step 2 (line 4 in Figure 2) finds the optimal basis function $h$, as well as its optimal coefficient $\beta$, that best fits the residual according to the least-squares (LS) criterion. Formally, let $h$ and $\beta$ denote the candidate basis function and its optimal coefficient. The LS error on training data is $LS(h;\beta) = \sum_{i=0}^{N}(y_i' - \beta h)^2$, where $y_i'$ is computed as Equation (5). The optimal coefficient of $h$ is estimated by solving the equation $\partial \sum_{i=1}^{N}(y_i' - \beta h)^2 / \partial \beta = 0$. Then, $\beta$ is computed as

$$\beta = \frac{\sum_{i=1}^{N} y_i' h(\mathbf{x}_i)}{\sum_{i=1}^{N} h(\mathbf{x}_i)}. \quad (8)$$

Finally, given its optimal coefficient $\beta$, the optimal LS loss of $h$ is

$$LS(h;\beta) = \sum_{i=1}^{N} y_i' \times y_i' - \frac{(\sum_{i=1}^{N} y_i' h(\mathbf{x}_i))^2}{\sum_{i=1}^{N} h^2(\mathbf{x}_i)}. \quad (9)$$

Step 3 (line 5 in Figure 2) updates the base model by adding the chosen optimal basis function with its optimal coefficient. As shown in Step 2, the optimal coefficient of each candidate basis function is computed when the basis function is evaluated. However, adding the basis function using its optimal efficient is prone to overfitting. We thus add a shrinkage coefficient $0 < \upsilon < 1$ – the fraction of the optimal line step taken. The update equation is thus rewritten in line 5 in Figure 2.

Notice that if the background model contains all the input features in $\mathbf{x}$, then LambdaBoost does not add any new features but adjust the weights of existing features. If the background model does not contain all of the input features, then LambdaBoost can be viewed as a feature selection method, similar to Collins (2000), where at each iteration the feature that has the largest impact on reducing training loss is selected and added to the background model. In either case, LambdaBoost adapts the background model by adding a model whose form is a (weighted) linear combination of input features. The property of linearity makes LambdaBoost robust and less likely to overfit in Web search applications. But this also limits the adaptation capacity. A simple method that allows us to go beyond linear adaptation is to define $h$ as

```
1  Set F_0(x) to be the background ranking model
2  for m = 1 to M do
3     compute residuals according to Equation (5)
4     create a L-terminal node tree, h_m ≡ {R_lm}_{l=1...L}
5     for l = 1 to L do
6        compute the optimal β_lm according to Equation (10),
         based on approximate Newton step.
7     F_m(x) = F_{m-1}(x) + υ ∑_{l=1...L} β_lm 1(x ∈ R_lm)
```

**Figure 3.** The LambdaSMART algorithm for model adaptation.

nonlinear terms of the input features, such as regression trees in LambdaSMART.

### 4.2 The LambdaSMART Algorithm

LambdaSMART was originally proposed in Wu et al. (2008). It is built on MART (Friedman, 2001) but uses the λ-function (Burges et a., 2006) to compute gradients. The algorithm is summarized in Figure 3. Similar to LambdaBoost, it takes $M$ rounds, and at each boosting iteration, it adapts the background model to training data using the 3-step procedure. Step 1 (line 3 in Figure 3) has been described.

Step 2 (lines 4 to 6) searches for the optimal basis function $h$ to best fit the residual. Unlike LambdaBoost where there are a finite number of candidate basis functions, the function space of regression trees is infinite. We define $h$ as a regression tree with $L$ terminal nodes. In line 4, a regression tree is built using Mean Square Error to determine the best split at any node in the tree. The value associated with a leaf (i.e., terminal node) of the trained tree is computed first as the residual (computed via λ-function) for the training samples that land at that leaf. Then, since each leaf corresponds to a different mean, a one-dimensional Newton-Raphson line step is computed for each leaf (lines 5 and 6). These line steps may be simply computed as the derivatives of the LambdaRank gradients w.r.t. the model scores $s_i$. Formally, the value of the $l$-th leaf, $β_{ml}$, is computed as

$$β_{ml} = \frac{\sum_{x \in R_{lm}} y_i'}{\sum_{x \in R_{lm}} w_i}, \qquad (10)$$

where $y_i'$ is the residual of training sample $i$, computed in Equation (5), and $w_i$ is the derivative of $y_i'$, i.e., $w_i = \partial y_i' / \partial F(x_i)$.

In Step 3 (line 7), the regression tree is added to the current base model, weighted by the shrinkage coefficient $0 < υ < 1$.

Notice that since a regression tree can be viewed as a complex feature that combines multiple input features, LambdaSMART can be used as a feature generation method. LambdaSMART is arguably more powerful than LambdaBoost in that it introduces new complex features and thus adjusts not only the parameters but also the structure of the background model[1]. However, one problem of trees is their high variance. Often a small change in the data can result in a very different series of splits. As a result, tree-based ranking models are much less robust to noise, as we will show in our experiments. In addition to the use of shrinkage coefficient $0 < υ < 1$, which is a form of model regularization according to Hastie, et al., (2001), we will explore in Section 5.3 other methods of improving the model robustness, including randomization and using shallow trees.

## 5 Experiments

### 5.1 The Data

We evaluated the ranking model adaptation methods on two Web search domains, namely (1) a name query domain, which consists of only person name queries, and (2) a Korean query domain, which consists of queries that users submitted to the Korean market.

For each domain, we used two in-domain data sets that contain queries sampled respectively from the query log of a commercial Web search engine that were collected in two non-overlapping periods of time. We used the more recent one as *open test* set, and split the other into three non-overlapping data sets, namely training, validation and *closed test* sets, respectively. This setting provides a good simulation to the realistic Web search scenario, where the rankers in use are usually trained on early collected data, and thus helps us investigate the robustness of these model adaptation methods.

The statistics of the data sets used in our person name domain adaptation experiments are shown in Table 1. The names query set serves as the adaptation domains, and Web-1 as the background domain. Since Web-1 is used to train a background ranker, we did not split it to train/valid/test sets. We used 416 input features in these experiments.

For cross-domain adaptation experiments from non-Korean to Korean markets, Korean data serves as the adaptation domain, and English, Chinese, and Japanese data sets as the background domain. Again, we did not split the data sets in the background domain to train/valid/test sets. The statistics of these data sets are shown in Table 2. We used 425 input features in these experiments.

---

[1] Note that in a sense our proposed LambdaBoost algorithm is the same as LambdaSMART, but using a single feature at each iteration, rather than a tree. In particular, they share the trick of using the Lambda gradients to learn NDCG.

| Coll. | Description | # qry. | # url/qry |
|---|---|---|---|
| **Web-1** | *Background training data* | 31555 | 134 |
| **Names-1-Train** | *In-domain training data (adaptation data)* | 5752 | 85 |
| **Names-1-Valid** | *In-domain validation data* | 158 | 154 |
| **Names-1-Test** | *Closed test data* | 318 | 153 |
| **Names-2-Test** | *Open test data* | 4370 | 84 |

**Table 1.** Data sets in the names query domain experiments, where # qry is number of queries, and # url/qry is number of documents per query.

| Coll. | Description | # qry. | # url/qry |
|---|---|---|---|
| **Web-En** | *Background English training data* | 6167 | 198 |
| **Web-Ja** | *Background Japanese training data* | 45012 | 58 |
| **Web-Cn** | *Background Chinese training data* | 32827 | 72 |
| **Kokr-1-Train** | *In-domain Korean training data (adaptation data)* | 3724 | 64 |
| **Kokr-1-Valid** | *In-domain validation data* | 334 | 130 |
| **Kokr-1-Test** | *Korean closed test data* | 372 | 126 |
| **Kokr-2-Test** | *Korean open test data* | 871 | 171 |

**Table 2.** Data sets in the Korean domain experiments.

| # | Models | NDCG@1 | NDCG@3 | NDCG@10 | AveNDCG |
|---|---|---|---|---|---|
| 1 | Back. | 0.4575 | 0.4952 | 0.5446 | 0.5092 |
| 2 | In-domain | 0.4921 | 0.5296 | 0.5774 | 0.5433 |
| 3 | 2W-Interp. | 0.4745 | 0.5254 | 0.5747 | 0.5391 |
| 4 | 3W-Interp. | 0.4829 | 0.5333 | 0.5814 | 0.5454 |
| 5 | λ-Boost | 0.4706 | 0.5011 | 0.5569 | 0.5192 |
| 6 | λ-SMART | 0.5042 | 0.5449 | 0.5951 | **0.5623** |

**Table 3.** Close test results on Names-1-Test.

| # | Models | NDCG@1 | NDCG@3 | NDCG@10 | AveNDCG |
|---|---|---|---|---|---|
| 1 | Back. | 0.5472 | 0.5347 | 0.5731 | 0.5510 |
| 2 | In-domain | 0.5216 | 0.5266 | 0.5789 | 0.5472 |
| 3 | 2W-Interp. | 0.5452 | 0.5414 | 0.5891 | 0.5604 |
| 4 | 3W-Interp. | 0.5474 | 0.5470 | 0.5951 | **0.5661** |
| 5 | λ-Boost | 0.5269 | 0.5233 | 0.5716 | 0.5428 |
| 6 | λ-SMART | 0.5200 | 0.5331 | 0.5875 | 0.5538 |

**Table 4.** Open test results on Names-2-Test.

In each domain, the in-domain training data is used to train in-domain rankers, and the background data for background rankers. Validation data is used to learn the best training parameters of the boosting algorithms, i.e., *M*, the total number of boosting iterations, $\upsilon$, the shrinkage coefficient, and *L*, the number of leaf nodes for each regression tree (*L*=1 in LambdaBoost). Model performance is evaluated on the closed/open test sets.

All data sets contain samples labeled on a 5-level relevance scale, 0 to 4, with 4 as most relevant and 0 as irrelevant. The performance of rankers is measured through NDCG evaluated against closed/open test sets. We report NDCG scores at positions 1, 3 and 10, and the averaged NDCG score (Ave-NDCG), the arithmetic mean of the NDCG scores at 1 to 10. Significance test (i.e., t-test) was also employed.

### 5.2 Model Adaptation Results

This section reports the results on two adaptation experiments. The first uses a large set of Web data, Web-1, as background domain and uses the name query data set as adaptation data. The results are summarized in Tables 3 and 4. We compared the three model adaptation methods against two baselines: (1) the background ranker (Row 1 in Tables 3 and 4), a 2-layer LambdaRank model with 15 hidden nodes and a learning rate of $10^{-5}$ trained on Web-1; and (2) the In-domain Ranker (Row 2), a 2-layer LambdaRank model with 10 hidden nodes and a learning rate of $10^{-5}$ trained on Names-1-Train. We built two interpolated rankers. The 2-way interpolated ranker (Row 3) is a linear combination of the two baseline rankers, where the interpolation weights were optimized on Names-1-Valid. To build the 3-way interpolated ranker (Row 4), we linearly interpolated three rankers. In addition to the two baseline rankers, the third ranker is trained on an augmented training data, which was created using the kNN method described in Section 3.

In LambdaBoost (Row 5) and LambdaSMART (Row 6), we adapted the background ranker to name queries by boosting the background ranker with Names-1-Train. We trained LambdaBoost with the setting $M = 500$, $\upsilon = 0.5$, optimized on Names-1-Valid. Since the background ranker uses all of the 416 input features, in each boosting iteration, LambdaBoost in fact selects one existing feature in the background ranker and adjusts its weight. We trained LambdaSMART with $M = 500$, $L = 20$, $\upsilon = 0.5$, optimized on Names-1-Valid.

We see that the results on the closed test set (Table 3) are quite different from the results on the open test set (Table 4). The in-domain ranker outperforms the background ranker on the closed test set, but underperforms significantly the background ranker on the open test set. The interpretation is that the training set and the closed test set are sampled from the same data set and are very similar, but the open test set is a very different data set, as described in Section 5.1. Similarly, on the closed test set, LambdaSMART outperforms LambdaBoost with a big margin due to its superior

| # | Ranker | NDCG@1 | NDCG@3 | NDCG@10 | AveNDCG |
|---|---|---|---|---|---|
| 1 | Back. (En) | 0.5371 | 0.5413 | 0.5873 | 0.5616 |
| 2 | Back. (Ja) | 0.5640 | 0.5684 | 0.6027 | 0.5808 |
| 3 | Back. (Cn) | 0.4966 | 0.5105 | 0.5761 | 0.5393 |
| 4 | In-domain | 0.5927 | 0.5824 | 0.6291 | **0.6055** |

Table 5. Close test results of baseline rankers, tested on Kokr-1-Test.

| # | Ranker | NDCG@1 | NDCG@3 | NDCG@10 | AveNDCG |
|---|---|---|---|---|---|
| 1 | Back. (En) | 0.4991 | 0.5242 | 0.5397 | 0.5278 |
| 2 | Back. (Ja) | 0.5052 | 0.5092 | 0.5377 | 0.5194 |
| 3 | Back. (Cn) | 0.4779 | 0.4855 | 0.5114 | 0.4942 |
| 4 | In-domain | 0.5164 | 0.5295 | 0.5675 | **0.5430** |

Table 6. Open test results of baseline rankers, tested on Kokr-2-Test.

| # | Ranker | NDCG@1 | NDCG@3 | NDCG@10 | AveNDCG |
|---|---|---|---|---|---|
| 1 | Interp. (En) | 0.5954 | 0.5893 | 0.6335 | **0.6088** |
| 2 | Interp. (Ja) | 0.6047 | 0.5898 | 0.6339 | 0.6116 |
| 3 | Interp. (Cn) | 0.5812 | 0.5807 | 0.6268 | 0.6024 |
| 4 | 4W-Interp. | 0.5878 | 0.5870 | 0.6289 | 0.6054 |

Table 7. Close test results of interpolated rankers, on Kokr-1-Test.

| # | Ranker | NDCG@1 | NDCG@3 | NDCG@10 | AveNDCG |
|---|---|---|---|---|---|
| 1 | Interp. (En) | 0.5178 | 0.5369 | 0.5768 | 0.5500 |
| 2 | Interp. (Ja) | 0.5274 | 0.5416 | 0.5788 | 0.5531 |
| 3 | Interp. (Cn) | 0.5224 | 0.5339 | 0.5766 | 0.5487 |
| 4 | 4W-Interp. | 0.5278 | 0.5414 | 0.5823 | **0.5549** |

Table 8. Open test results of interpolated rankers, on Kokr-2-Test.

| # | Ranker | NDCG@1 | NDCG@3 | NDCG@10 | AveNDCG |
|---|---|---|---|---|---|
| 1 | λ-Boost (En) | 0.5757 | 0.5716 | 0.6197 | 0.5935 |
| 2 | λ-Boost (Ja) | 0.5801 | 0.5807 | 0.6225 | **0.5982** |
| 3 | λ-Boost (Cn) | 0.5731 | 0.5793 | 0.6226 | 0.5972 |

Table 9. Close test results of λ-Boost rankers, on Kokr-1-Test.

| # | Ranker | NDCG@1 | NDCG@3 | NDCG@10 | AveNDCG |
|---|---|---|---|---|---|
| 1 | λ-Boost (En) | 0.4960 | 0.5203 | 0.5486 | 0.5281 |
| 2 | λ-Boost (Ja) | 0.5090 | 0.5167 | 0.5374 | 0.5233 |
| 3 | λ-Boost (Cn) | 0.5177 | 0.5324 | 0.5673 | **0.5439** |

Table 10. Open test results of λ-Boost rankers, on Kokr-2-Test.

| # | Ranker | NDCG@1 | NDCG@3 | NDCG@10 | AveNDCG |
|---|---|---|---|---|---|
| 1 | λ-SMART (En) | 0.6096 | 0.6057 | 0.6454 | **0.6238** |
| 2 | λ-SMART (Ja) | 0.6014 | 0.5966 | 0.6385 | 0.6172 |
| 3 | λ-SMART (Cn) | 0.5955 | 0.6095 | 0.6415 | 0.6209 |

Table 11. Close test results of λ-SMART rankers, on Kokr-1-Test.

| # | Ranker | NDCG@1 | NDCG@3 | NDCG@10 | AveNDCG |
|---|---|---|---|---|---|
| 1 | λ-SMART (En) | 0.5177 | 0.5297 | 0.5563 | 0.5391 |
| 2 | λ-SMART (Ja) | 0.5205 | 0.5317 | 0.5522 | 0.5368 |
| 3 | λ-SMART (Cn) | 0.5198 | 0.5305 | 0.5644 | **0.5410** |

Table 12. Open test results of λ-SMART rankers, on Kokr-2-Test.

adaptation capacity; but on the open test set their performance difference is much smaller due to the instability of the trees in LambdaSMART, as we will investigate in detail later. Interestingly, model interpolation, though simple, leads to the two best rankers on the open test set. In particular, the 3-way interpolated ranker outperforms the two baseline rankers significantly (i.e., $p$-value < 0.05 according to t-test) on both the open and closed test sets.

The second adaptation experiment involves data sets from several languages (Table 2). 2-layer LambdaRank baseline rankers were first built from Korean, English, Japanese, and Chinese training data and tested on Korean test sets (Tables 5 and 6). These baseline rankers then serve as in-domain ranker and background rankers for model adaptation. For model interpolation (Tables 7 and 8), Rows 1 to 4 are three 2-way interpolated rankers built by linearly interpolating each of the three background rankers with the in-domain ranker, respectively. Row 4 is a 4-way interpolated ranker built by interpolating the in-domain ranker with the three background rankers. For LambdaBoost (Tables 9 and 10) and LambdaSMART (Tables 11 and 12), we used the same parameter settings as those in the name query experiments, and adapted the three background rankers, to the Korean training data, Kokr-1-Train.

The results in Tables 7 to 12 confirm what we learned in the name query experiments. There are three main conclusions. (1) Model interpolation is an effective method of ranking model adaptation. E.g., the 4-way interpolated ranker outperforms other ranker significantly. (2) LambdaSMART is the best performer on the closed test set, but its performance drops significantly on the open test set due to the instability of trees. (3) LambdaBoost does not use trees. So its modeling capacity is weaker than LambdaSMART (e.g., it always underperforms LambdaSMART significantly on the closed test sets), but it is more robust due to its linearity (e.g., it performs similarly to LambdaSMART on the open test set).

## 5.3 Robustness of Boosting Algorithms

This section investigates the robustness issue of the boosting algorithms in more detail. We compared LambdaSMART with different values of $L$ (i.e., the number of leaf nodes), and with and without randomization. Our assumptions are (1) allowing more leaf nodes would lead to deeper trees, and as a result, would make the resulting ranking models less robust; and (2) injecting randomness into the basis function (i.e. regression tree) estimation procedure would improve the robustness of the trained models (Breiman, 2001; Friedman, 1999). In LambdaSMART, the randomness can be injected at different levels of tree construction. We found that the most effective method is to introduce the randomness at the node level (in Step 4 in Figure 3). Before each node split, a subsample of the training data and a subsample of the features are drawn randomly. (The sample rate is 0.7). Then, the two randomly selected subsamples, instead of the full samples, are used to determine the best split.

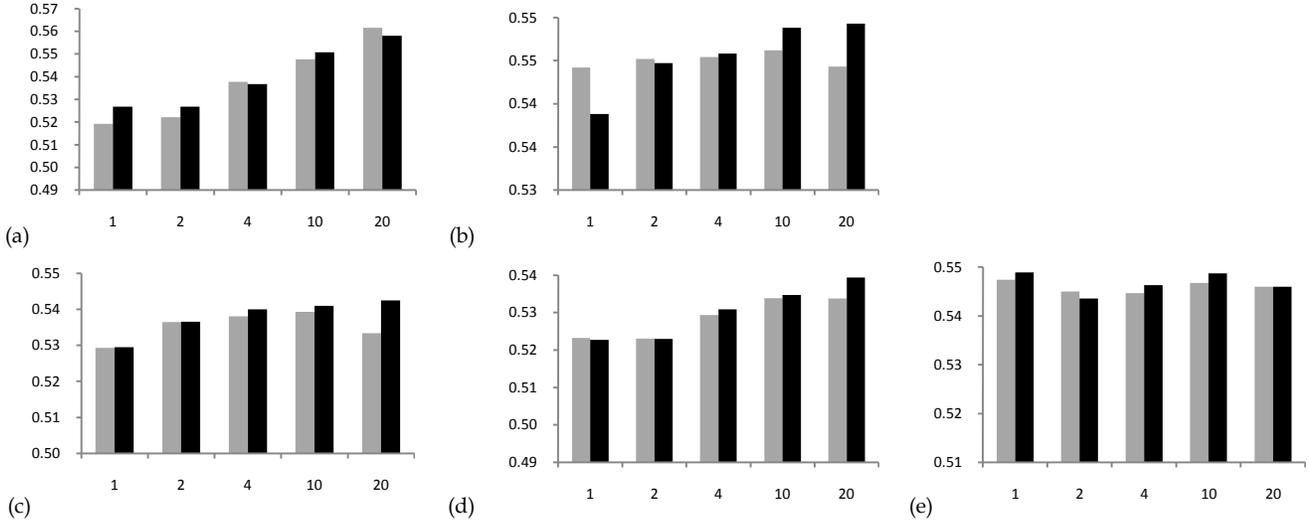

**Figure 4.** AveNDCG results (y-axis) of LambdaSMART with different values of L (x-axis), where L=1 is LambdaBoost; (a) and (b) are the results on closed and open tests using Names-1-Train as adaptation data, respectively; (d), (e) and (f) are the results on the Korean open test set, using background models trained on Web-En, Web-Ja, and Web-Cn data sets, respectively.

We first performed the experiments on name queries. The results on the closed and open test sets are shown in Figures 4 (a) and 4 (b), respectively. The results are consistent with our assumptions. There are three main observations. First, the gray bars in Figures 4 (a) and 4 (b) (boosting without randomization) show that on the closed test set, as expected, NDCG increases with the value of $L$, but the correlation does not hold on the open test set. Second, the black bars in these figures (boosting with randomization) show that in both closed and open test sets, NDCG increases with the value of $L$. Finally, comparing the gray bars with their corresponding black bars, we see that randomization consistently improves NDCG on the open test set, with a larger margin of gain for the boosting algorithms with deeper trees ($L > 5$).

These results are very encouraging. Randomization seems to work like a charm. Unfortunately, it does not work well enough to help the boosting algorithm beat model interpolation on the open test sets. Notice that all the LambdaSMART results reported in Section 5.2 use randomization with the same sampling rate of 0.7. We repeated the comparison in the cross-domain adaptation experiments. As shown in Figure 4, results in 4 (c) and 4 (d) are consistent with those on names queries in 4 (b). Results in 4 (f) show a visible performance drop from LambdaBoost to LambdaSMART with $L = 2$, indicating again the instability of trees.

## 6 Conclusions and Future Work

In this paper, we extend two classes of model adaptation methods (i.e., model interpolation and error-driven learning), which have been well studied in statistical language modeling for speech and natural language applications (e.g., Bacchiani et al., 2004; Bellegarda, 2004; Gao et al., 2006), to ranking models for Web search applications.

We have evaluated our methods on two adaptation experiments over a wide variety of datasets where the in-domain datasets bear different levels of similarities to their background datasets. We reach different conclusions from the results of the open and close tests, respectively. Our open test results show that in the cases where the in-domain data is dramatically different from the background data, model interpolation is very robust and outperforms the baseline and the error-driven learning methods significantly; whereas our close test results show that in the cases where the in-domain data is similar to the background data, the tree-based boosting algorithm (i.e. LambdaSMART) is the best performer, and achieves a significant improvement over the baselines. We also show that these different conclusions are largely due to the instability of the use of trees in the boosting algorithm. We thus explore several methods of improving the robustness of the algorithm, such as randomization, regularization, using shallow trees, with limited success. Of course, our experiments, described in Section 5.3, only scratch the surface of what is possible. Robustness deserves more investigation and forms one area of our future work.

Another family of model adaptation methods that we have not studied in this paper is transfer learning, which has been well-studied in the machine learning community (e.g., Caruana, 1997; Marx et al., 2008). We leave it to future work.

To solve the issue of inadequate training data, in addition to model adaptation, researchers have also been exploring the use of implicit user feedback data (extracted from log files) for ranking model training (e.g., Joachims et al., 2005; Radlinski et al., 2008). Although such data is very noisy, it is of a much larger amount and is cheaper to obtain than human-labeled data. It will be interesting to apply the model adaptation methods described in this paper to adapt a ranker which is trained on a large amount of automatically extracted data to a relatively small amount of human-labeled data.

## Acknowledgments

This work was done while Yi Su was visiting Microsoft Research, Redmond. We thank Steven Yao's group at Microsoft Bing Search for their help with the experiments.